\documentclass{article}
\usepackage{spconf,amsmath,graphicx}
\usepackage{amsmath,amssymb}
\usepackage{xcolor}
\usepackage{tikz}
\usetikzlibrary{arrows.meta,positioning,calc,fit,backgrounds,shapes.geometric}
\definecolor{AStarGreen}{RGB}{88, 255, 100} 

\usepackage[
  colorlinks=true,
  linkcolor=AStarGreen,   
  citecolor=AStarGreen,   
  urlcolor=AStarGreen     
]{hyperref}
\usepackage{url}

\title{Is Haar Enough? Exploring Symlets and Coiflets for Wavelet Convolution Layers}

\name{Md Rifat Ur Rahman}
\address{Bangladesh Univeristy of Engineering and Technology}

\begin{document}
%
\maketitle
\begin{abstract}
Wavelet convolution layers have recently emerged as an efficient mechanism for enlarging receptive fields through multiresolution analysis, but prior work has fixed the wavelet basis to Haar or Daubechies at a chosen decomposition depth, leaving open whether a different basis can shift the underlying efficiency frontier. We identify and characterize a previously unexplored trade-off in this setting: bases with stronger approximation properties (longer filters) can reduce the decomposition depth required for competitive accuracy, yielding a net reduction in parameters and FLOPs despite higher per-level transform cost. We formalize this as an $F$-vs.-$L$ trade-off (filter length vs.\ decomposition levels) and study it systematically across Haar, Daubechies, Symlets, and Coiflets under controlled architectures and budgets. On image classification (CIFAR-10, ImageNet-1K) and semantic segmentation (Cityscapes), Coiflet-based wavelet convolutions match Haar at deeper levels with approximately 32\% fewer additional parameters and 33\% fewer additional FLOPs, providing a concrete and actionable design choice for practitioners building wavelet-based architectures. Code will be released.
\end{abstract}
\section{Introduction}
\label{sec:intro}

Convolutional neural networks (CNNs) have been the default backbone for many computer vision tasks since AlexNet~\cite{krizhevsky2012imagenet}, with strong subsequent designs such as VGG~\cite{simonyan2015verydeep} and ResNet~\cite{he2016deep}. 
A key strength of CNNs is their inductive bias toward local processing: convolution and pooling efficiently capture fine-grained, high-frequency patterns (e.g., edges and textures). 
However, modeling long-range interactions typically requires architectural choices that enlarge the effective receptive field, such as deeper networks, multi-stage hierarchies, dilation, or large kernels, which can increase compute and/or parameters.

Vision Transformers (ViTs)~\cite{dosovitskiy2021image} address long-range interaction more directly by using self-attention~\cite{vaswani2017attention}, enabling global token mixing. 
Nevertheless, attention-based models often incur higher computational and memory costs, motivating a large body of work on efficient backbones and token mixers. 
On the CNN side, lightweight architectures such as MobileNet~\cite{howard2017mobilenets} and EfficientNet~\cite{tan2019efficientnet} reduce compute via depthwise separable and inverted bottleneck designs, but the core local nature of convolution remains. 
On the token-mixing side, large-kernel designs aim to approximate long-range interactions with convolutional operators; for example, Visual Attention Network (VAN) introduces Large Kernel Attention (LKA) as a lightweight mechanism for long-range dependency modeling~\cite{guo2022van}. 

Frequency-domain token mixers provide another direction: GFNet replaces attention with Fourier-domain global filtering via FFT and inverse FFT~\cite{rao2021gfnet}, AFFNet proposes adaptive frequency filters as efficient token mixers~\cite{huang2023aff}, and Fourier Neural Operators (FNO) learn global convolution operators in the Fourier domain for operator learning and related vision tasks~\cite{li2021fno}. 
While these Fourier-based approaches provide global mixing efficiently, Fourier-domain filtering is inherently global in spatial support, and in practice is often paired with spatial/local processing (e.g., convolutional stems or hybrid blocks) to retain fine-grained locality.

In parallel, wavelet transforms offer a principled multiresolution representation that jointly captures spatial and frequency information~\cite{mallat1989theory}. 
In a 2D discrete wavelet transform (2D-DWT), each level decomposes a feature map into four sub-bands: a low-frequency approximation band (LL) and three high-frequency detail bands (LH/HL/HH). 
By iteratively decomposing the LL band, multilevel DWT provides progressively coarser representations while preserving global structure at reduced spatial resolution. 
This motivates wavelet-based architectures such as WaveMix~\cite{jeevan2022wavemix} and Wave-ViT~\cite{yao2022wavevit}, as well as wavelet-inspired attention variants (e.g., multiscale wavelet attention)~\cite{nekoozadeh2023mwa}.

Recently, WTConv introduced a wavelet convolution layer as a plug-in replacement for depthwise convolution to obtain large receptive fields with favorable scaling~\cite{finder2024wtconv}. 
WTConv applies multilevel 2D-DWT, performs depthwise convolution in the wavelet domain across sub-bands, and reconstructs via inverse DWT. 
The original formulation primarily uses the Haar basis (db1) and reports limited gains when switching to db2/db3 \emph{at a fixed decomposition depth}~\cite{finder2024wtconv}. 
We argue that this fixed-depth comparison conceals the actual design space: because longer-filter bases provide stronger approximation in the low-frequency (LL) branch, the depth required to capture global context is itself a function of the basis. This raises a natural and, to our knowledge, previously unstudied question: \emph{when the basis and the number of decomposition levels are co-optimized, can alternative bases provide a strictly better accuracy--complexity frontier than Haar?}

In this paper, we revisit wavelet basis selection for wavelet convolution layers under this lens. 
Our choice of bases is hypothesis-driven, not exhaustive: we select Symlets and Coiflets as widely used orthogonal wavelet families beyond Haar/Daubechies~\cite{daubechies1992ten} because they target the two properties most relevant to the depth question. 
Symlets are near-symmetric, which reduces phase distortion and can improve edge alignment across scales. 
Coiflets have additional vanishing moments on the scaling function and are known to yield smoother LL representations, which we hypothesize should reduce the depth needed to summarize global context. 
Both predictions are testable in our setup, and our experiments confirm them for Coiflets.

We evaluate basis choices on image classification (CIFAR-10, ImageNet-1K) and semantic segmentation (Cityscapes) under controlled architectures, and report accuracy alongside model parameters and FLOPs.

Our main contributions are:
\begin{itemize}
    \item We identify a previously unexplored \emph{filter-length vs.\ decomposition-depth} ($F$-vs.-$L$) trade-off in wavelet convolution layers. Prior work fixed the depth and varied only the basis; we show that allowing both to vary changes the qualitative conclusion of which basis is preferable.
    \item We give a hypothesis-driven analysis of Symlets and Coiflets as candidates for this trade-off, motivated by their symmetry and approximation-order properties respectively, and confirm the Coiflet hypothesis empirically.
    \item Across classification and segmentation, we show that Coiflet-based WTConv matches deeper Haar configurations with approximately 32\% fewer additional parameters and 33\% fewer additional FLOPs, providing a concrete and actionable design choice for wavelet-based architectures.
\end{itemize}
\section{Methodology}
\label{sec:method}

We study whether the \emph{wavelet basis} used inside wavelet convolution layers can be improved beyond commonly used Haar/Daubechies choices.
Our operator follows a transform--process--reconstruct pipeline: a feature map is decomposed by a multi-level 2D discrete wavelet transform (2D-DWT), filtered in the wavelet domain using lightweight depthwise convolution, and reconstructed by inverse DWT (IDWT).
The two main knobs in this work are (i) the chosen wavelet basis (fixed analysis/synthesis filters) and (ii) the number of decomposition levels $L$.
We evaluate the resulting trade-off in accuracy vs.\ computational complexity (FLOPs) and parameters.

\subsection{Wavelet Transform}
\label{sec:wt}

Let the input feature map be
\begin{equation}
\label{eq:x_def}
X \in \mathbb{R}^{B \times C \times H \times W}.
\end{equation}
A one-level 2D-DWT decomposes $X$ into four critically-sampled sub-bands:
\begin{equation}
\label{eq:dwt_outputs}
\left(X_{LL}, X_{LH}, X_{HL}, X_{HH}\right) = \mathrm{DWT}(X),
\end{equation}
with each sub-band having half spatial resolution:

Where, \(X_{\bullet} \in \mathbb{R}^{B \times C \times \frac{H}{2} \times \frac{W}{2}}\). For implementation, we pack the four sub-bands along channels:
\begin{equation}
\label{eq:pack}
Y^{(1)} = \mathrm{Pack}\!\left(X_{LL}^{(1)}, X_{LH}^{(1)}, X_{HL}^{(1)}, X_{HH}^{(1)}\right),
\end{equation}

Where, \(Y^{(1)} \in \mathbb{R}^{B \times 4C \times \frac{H}{2} \times \frac{W}{2}}\). A multi-level pyramid is obtained by recursively decomposing only the approximation band:
\begin{equation}
\label{eq:multilevel}
\left(X_{LL}^{(i)}, X_{LH}^{(i)}, X_{HL}^{(i)}, X_{HH}^{(i)}\right)=\mathrm{DWT}\!\left(X_{LL}^{(i-1)}\right),
\quad i=2,\dots,L.
\end{equation}

\begin{figure}[h]
    \centering
    \includegraphics[width=1.0\linewidth]{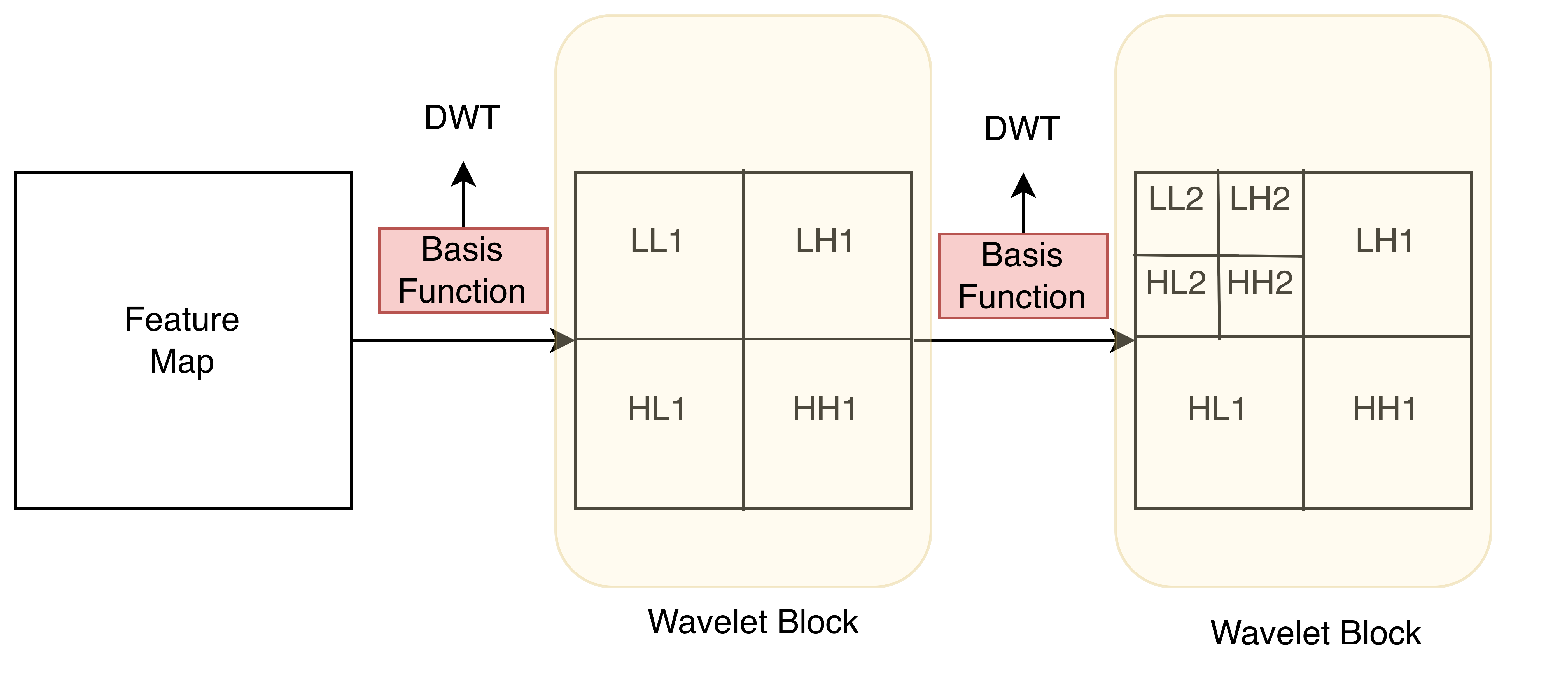}
    \caption{Two-level 2D-DWT decomposition used in our wavelet convolution layer. The first level decomposes the feature map into $(LL_1, LH_1, HL_1, HH_1)$, and the second level further decomposes only $LL_1$ into $(LL_2, LH_2, HL_2, HH_2)$. We evaluate alternative wavelet bases by replacing the fixed analysis/synthesis filter banks.}
\label{fig:wavelet_block}
\end{figure}

Reconstruction uses the inverse transform:
\begin{equation}
\label{eq:idwt_def}
X_{LL}^{(i-1)}=\mathrm{IDWT}\!\left(X_{LL}^{(i)}, X_{LH}^{(i)}, X_{HL}^{(i)}, X_{HH}^{(i)}\right),
\end{equation}
and is applied level-by-level from $i=L$ down to $i=1$.

\subsection{Wavelet Basis Selection}
\label{sec:basis}

\begin{table*}[t]
\centering
\caption{1D analysis filter taps used in our 2D-DWT. The 2D sub-band filters are formed via outer products in Eq.~\eqref{eq:outer_filters}. Coefficients are taken from a standard wavelet library for reproducibility~\cite{pywavelets}.}
\label{tab:taps}
\setlength{\tabcolsep}{7pt}
\renewcommand{\arraystretch}{1.05}
\footnotesize
\begin{tabular}{l l p{0.72\textwidth}}
\hline
Family & Filter & Coefficients \\
\hline
Haar (db1) & $h_0$ &
$[\tfrac{1}{\sqrt{2}},\; \tfrac{1}{\sqrt{2}}]$ \\
          & $h_1$ &
$[-\tfrac{1}{\sqrt{2}},\; \tfrac{1}{\sqrt{2}}]$ \\
\hline
Daubechies (db2) & $h_0$ &
$[-0.12940952,\; 0.22414387,\; 0.83651630,\; 0.48296291]$ \\
                 & $h_1$ &
$[-0.48296291,\; 0.83651630,\; -0.22414387,\; -0.12940952]$ \\
\hline
Daubechies (db3) & $h_0$ &
$[0.03522629,\; -0.08544127,\; -0.13501102,\; 0.45987750,\; 0.80689151,\; 0.33267055]$ \\
                 & $h_1$ &
$[-0.33267055,\; 0.80689151,\; -0.45987750,\; -0.13501102,\; 0.08544127,\; 0.03522629]$ \\
\hline
Symlet (sym2) & $h_0$ &
$[-0.12940952,\; 0.22414387,\; 0.83651630,\; 0.48296291]$ \\
              & $h_1$ &
$[-0.48296291,\; 0.83651630,\; -0.22414387,\; -0.12940952]$ \\
\hline
Symlet (sym3) & $h_0$ &
$[0.03522629,\; -0.08544127,\; -0.13501102,\; 0.45987750,\; 0.80689151,\; 0.33267055]$ \\
              & $h_1$ &
$[-0.33267055,\; 0.80689151,\; -0.45987750,\; -0.13501102,\; 0.08544127,\; 0.03522629]$ \\
\hline
Coiflet (coif1) & $h_0$ &
$[-0.01565573,\; -0.07273262,\; 0.38486485,\; 0.85257202,\; 0.33789766,\; -0.07273262]$ \\
                & $h_1$ &
$[0.07273262,\; 0.33789766,\; -0.85257202,\; 0.38486485,\; 0.07273262,\; -0.01565573]$ \\
\hline
\end{tabular}
\end{table*}

A wavelet basis is defined by fixed 1D analysis filters: a low-pass filter $h_0$ and a high-pass filter $h_1$ (with matching synthesis filters used in IDWT).
For reproducibility, we report the exact analysis taps used in this work in Table~\ref{tab:taps} (from a standard wavelet library~\cite{pywavelets}).

Because the 2D-DWT is separable, the four 2D analysis filters are formed by outer products of the 1D filters:
\begin{equation}
\label{eq:outer_filters}
f_{LL}=h_0\otimes h_0,\;\;
f_{LH}=h_0\otimes h_1,\;\;
f_{HL}=h_1\otimes h_0,\;\;
f_{HH}=h_1\otimes h_1,
\end{equation}
and each sub-band is produced by a depthwise convolution with stride 2 (critical sampling), consistent with common wavelet-convolution implementations.

\vspace{0.2em}
\noindent\textbf{Haar (db1).}
Haar provides the shortest support ($F=2$) and thus the lowest transform overhead per level.
It yields a coarse, piecewise-constant approximation in the LL band, making it a strong efficiency baseline.

\vspace{0.2em}
\noindent\textbf{Daubechies (db$N$).}
Daubechies wavelets are compactly supported orthogonal wavelets that increase the number of vanishing moments with minimal support length ($F=2N$).
This typically improves approximation quality and frequency separation relative to Haar, at the cost of longer filters and higher per-level transform overhead.
We include db2 and db3 as representative longer-support baselines (Table~\ref{tab:taps}).

\vspace{0.2em}
\noindent\textbf{Symlets (sym$N$).}
Symlets are orthogonal wavelets designed to be more symmetric than standard Daubechies filters while retaining compact support.
Greater symmetry is commonly associated with reduced phase distortion and improved edge alignment across scales.
We evaluate sym2 and sym3 (Table~\ref{tab:taps}) as symmetry-oriented alternatives within the same general family of compactly supported orthogonal wavelets.

\vspace{0.2em}
\noindent\textbf{Coiflets (coif$N$).}
Coiflets are compactly supported orthogonal wavelets with strong approximation properties; empirically they often yield smoother low-frequency (LL) representations and a cleaner separation between approximation and detail components.
This can reduce the need for deeper decompositions to capture global context.
We evaluate coif1 (Table~\ref{tab:taps}) as a representative Coiflet basis.

\vspace{0.2em}
\noindent\textbf{Basis--level trade-off.}
Given the taps in Table~\ref{tab:taps}, all 2D sub-band filters are obtained deterministically via Eq.~\eqref{eq:outer_filters}.
Basis selection changes the filter length $F$ (support) and therefore the per-level DWT/IDWT cost, while potentially reducing the required decomposition depth $L$ to reach a target accuracy.
Our experiments quantify this $F$--vs.--$L$ trade-off by reporting accuracy together with FLOPs and parameters across bases and decomposition levels.

\subsection{Wavelet Convolution Layer}
\label{sec:wtconv_layer}

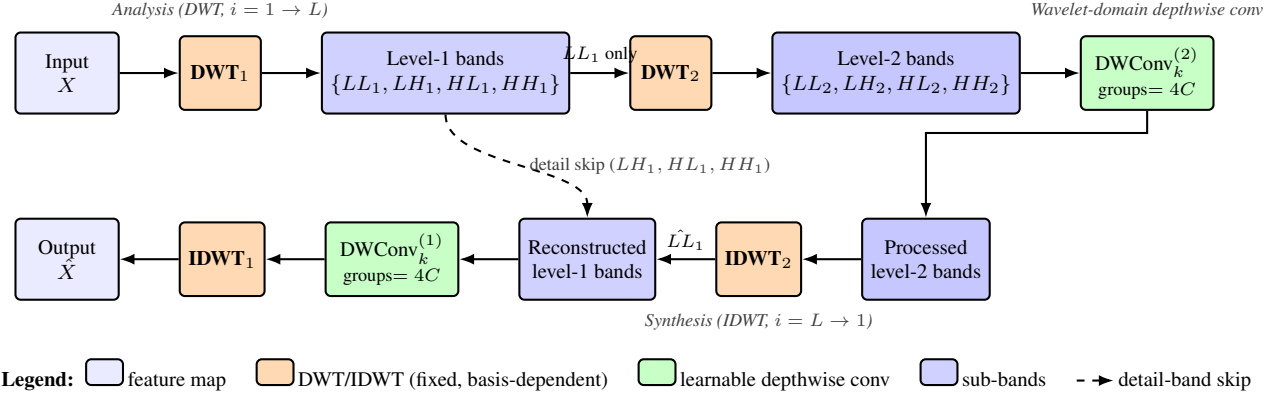
\begin{figure*}[t]
\centering
\resizebox{0.95\textwidth}{!}{%
\begin{tikzpicture}[
    >={Latex[length=2mm]},
    font=\footnotesize,
    node distance=6mm and 8mm,
    fmap/.style    ={draw, rounded corners=2pt, minimum width=14mm, minimum height=11mm, fill=blue!8,  thick, align=center},
    dwt/.style     ={draw, rounded corners=2pt, minimum width=11mm, minimum height=10mm, fill=orange!30, thick, align=center, font=\footnotesize\bfseries},
    idwt/.style    ={draw, rounded corners=2pt, minimum width=11mm, minimum height=10mm, fill=orange!30, thick, align=center, font=\footnotesize\bfseries},
    conv/.style    ={draw, rounded corners=2pt, minimum width=18mm, minimum height=10mm, fill=green!22,  thick, align=center, font=\footnotesize},
    bandgroup/.style={draw, rounded corners=2pt, minimum width=14mm, minimum height=11mm, thick, align=center, font=\footnotesize},
    lbl/.style     ={font=\footnotesize, align=center}
]

\node[fmap] (X) {Input\\$X$};
\node[dwt, right=of X] (DWT1) {DWT$_1$};

\node[bandgroup, right=of DWT1, fill=blue!18] (L1bands)
  {Level-1 bands\\$\{LL_1,LH_1,HL_1,HH_1\}$};

\node[dwt, right=of L1bands] (DWT2) {DWT$_2$};

\node[bandgroup, right=of DWT2, fill=blue!22] (L2bands)
  {Level-2 bands\\$\{LL_2,LH_2,HL_2,HH_2\}$};

\node[conv, right=of L2bands] (DW2) {DWConv$^{(2)}_k$\\\scriptsize groups$=4C$};

\draw[->, thick] (X) -- (DWT1);
\draw[->, thick] (DWT1) -- (L1bands);
\draw[->, thick] (L1bands) -- node[above, font=\scriptsize]{$LL_1$ only} (DWT2);
\draw[->, thick] (DWT2) -- (L2bands);
\draw[->, thick] (L2bands) -- (DW2);

\node[fmap, below=14mm of X] (Xhat) {Output\\$\hat{X}$};
\node[idwt, right=of Xhat] (IDWT1) {IDWT$_1$};
\node[conv, right=of IDWT1] (DW1) {DWConv$^{(1)}_k$\\\scriptsize groups$=4C$};
\node[bandgroup, right=of DW1, fill=blue!18] (L1rec)
  {Reconstructed\\level-1 bands};
\node[idwt, right=of L1rec] (IDWT2) {IDWT$_2$};
\node[bandgroup, right=of IDWT2, fill=blue!22] (L2proc)
  {Processed\\level-2 bands};

\draw[->, thick] (L2proc) -- (IDWT2);
\draw[->, thick] (IDWT2) -- node[above, font=\scriptsize]{$\hat{LL}_1$} (L1rec);
\draw[->, thick] (L1rec) -- (DW1);
\draw[->, thick] (DW1) -- (IDWT1);
\draw[->, thick] (IDWT1) -- (Xhat);

\draw[->, thick] (DW2.south) -- ++(0,-3mm) -| (L2proc.north);

\draw[->, thick, dashed] (L1bands.south) .. controls +(0,-7mm) and +(0,7mm) .. 
  node[midway, right=0.5mm, font=\scriptsize, gray!50!black]{detail skip ($LH_1,HL_1,HH_1$)}
  (L1rec.north);

\node[font=\scriptsize\itshape, gray!50!black, above=1mm of DWT1] {Analysis (DWT, $i=1\to L$)};
\node[font=\scriptsize\itshape, gray!50!black, above=1mm of DW2] {Wavelet-domain depthwise conv};
\node[font=\scriptsize\itshape, gray!50!black, below=1mm of IDWT2] {Synthesis (IDWT, $i=L\to 1$)};

\node[lbl, below=10mm of Xhat.south west, anchor=west, xshift=-3mm] (legend)
  {\textbf{Legend:}\hspace{1.5mm}%
   \tikz\node[fmap, minimum width=5mm, minimum height=3.5mm]{};\hspace{0.5mm}feature map\hspace{4mm}%
   \tikz\node[dwt, minimum width=5mm, minimum height=3.5mm]{};\hspace{0.5mm}DWT/IDWT (fixed, basis-dependent)\hspace{4mm}%
   \tikz\node[conv, minimum width=5mm, minimum height=3.5mm]{};\hspace{0.5mm}learnable depthwise conv\hspace{4mm}%
   \tikz\node[bandgroup, minimum width=5mm, minimum height=3.5mm, fill=blue!18]{};\hspace{0.5mm}sub-bands\hspace{4mm}%
   \tikz{\draw[->,dashed,thick](0,0)--(0.5,0);}\hspace{0.5mm}detail-band skip};

\end{tikzpicture}%
}
\caption{Two-level WTConv layer ($L=2$). \textbf{Top row (analysis):} the input $X$ is decomposed by DWT into four sub-bands; only the approximation band $LL_1$ is recursively decomposed at level 2. The deepest level's packed sub-bands are processed by a learnable depthwise convolution (groups $=4C$). \textbf{Bottom row (synthesis):} reconstruction proceeds level-by-level via IDWT from $i=L$ down to $i=1$, with detail bands $\{LH_1,HL_1,HH_1\}$ bypassing the deeper level via a skip path. The choice of wavelet basis affects only the fixed DWT/IDWT filters: longer-support bases (db$N$, sym$N$, coif$N$) raise per-level transform cost but, with stronger approximation, can reduce the required depth $L$ --- the $F$-vs.-$L$ trade-off central to this work.}
\label{fig:wtconv_layer}
\end{figure*}

We define the wavelet convolution layer as a sequence of operations:
\begin{equation}
\label{eq:pipeline}
X \xrightarrow{\mathrm{DWT}} \{Y^{(i)}\}_{i=1}^{L}
\xrightarrow{\mathrm{DWConv}} \{Z^{(i)}\}_{i=1}^{L}
\xrightarrow{\mathrm{IDWT}} \hat{X}.
\end{equation}

At each level $i$, we pack the four sub-bands into
\begin{equation}
\label{eq:yi_shape}
Y^{(i)} \in \mathbb{R}^{B \times 4C \times \frac{H}{2^i} \times \frac{W}{2^i}}.
\end{equation}
We then apply depthwise convolution in the wavelet domain:
\begin{equation}
\label{eq:dwconv}
Z^{(i)} = \mathrm{DWConv}_{k}^{(i)}\!\left(Y^{(i)}; W^{(i)}\right),
\quad \text{groups}=4C,
\end{equation}
where $k$ is the spatial kernel size and $W^{(i)}$ are learnable depthwise kernels.
Finally, we unpack $Z^{(i)}$ into the processed sub-bands and reconstruct level-by-level with IDWT from $i=L$ down to $i=1$ to obtain $\hat{X}$.

\subsection{Computational Cost}
\label{sec:cost}
We report parameters and FLOPs. Wavelet basis choice does not add learnable parameters because analysis/synthesis filters are fixed; trainable parameters arise from the backbone and wavelet-domain depthwise kernels $W^{(i)}$.

For depthwise convolution on $M$ channels with kernel $k\times k$ and spatial size $H'\times W'$, the multiply-add cost is approximately
\begin{equation}
\label{eq:dwflops}
\mathrm{FLOPs}(\mathrm{DWConv}) \approx 2\, M\, k^2\, H'\, W'.
\end{equation}
In our wavelet layer, $M=4C$ and $H'W'=\frac{HW}{4^i}$ at level $i$, so the total convolution cost across $L$ levels is
\begin{equation}
\label{eq:conv_total}
\mathrm{FLOPs}_{\mathrm{conv}} \approx \sum_{i=1}^{L} 2\cdot(4C)\cdot k^2 \cdot \frac{HW}{4^i}.
\end{equation}
\\
The transform overhead depends on the wavelet filter length $F$.
From Eq.~\eqref{eq:outer_filters}, each coefficient is a weighted sum over an $F\times F$ neighborhood (via separable 1D filtering), so longer-support bases increase per-level DWT/IDWT cost roughly linearly with $F$.
However, increasing $F$ can reduce the number of required levels $L$ for a target accuracy.
Our experiments quantify this $F$--vs.--$L$ trade-off by reporting accuracy together with FLOPs and parameters.
\section{Experiments}
\label{sec:exp}

\subsection{Experiments on Image Classification}
\label{sec:cls}

\subsubsection{Setup}
We evaluate on CIFAR-10~\cite{krizhevsky2009learning} and ImageNet-1K~\cite{deng2009imagenet}.
On CIFAR-10, all models are trained for 120 epochs under a controlled training budget.
On ImageNet-1K, we follow the standard ConvNeXt training recipe~\cite{liu2022convnet} (300 epochs, $224{\times}224$ resolution) and swap only the token mixer for fair comparison.
We report Top-1 accuracy (\%), number of parameters (M), and FLOPs (G) for a single forward pass.
For fairness, we keep the backbone capacity fixed (ConvNeXt-T~\cite{liu2022convnet}) when comparing different token mixers, and only swap the mixing module.

\subsubsection{Performance Comparison against Edge Models}
\label{sec:cls_edgecmp}
Table~\ref{tab:cls_edge} compares our coiflet-based WTConv variant against representative efficient backbones and frequency-domain token mixers:
ConvNeXt-T with GFNet~\cite{rao2021gfnet}, Fourier Neural Operator (FNO)~\cite{li2020fno}, and AFF~\cite{huang2023aff}.
We include ConvNeXt-T+WTConv (coif1) as our main model.

\begin{table}[!htbp]
\centering
\caption{CIFAR-10 classification comparison (120 epochs).}
\label{tab:cls_edge}
\setlength{\tabcolsep}{3pt}
\renewcommand{\arraystretch}{1.05}
\footnotesize
\begin{tabular}{l c c c}
\hline
Model & Top-1 (\%) $\uparrow$ & Params (M) $\downarrow$ & FLOPs (G) $\downarrow$ \\
\hline
ConvNeXt-T~\cite{liu2022convnet} & 81.0 & 28.6 & 4.5 \\
ConvNeXt-T + GFNet~\cite{rao2021gfnet} & 81.2 & 32.9 & 4.7 \\
ConvNeXt-T + FNO~\cite{li2020fno} & 81.3 & 31.8 & 4.9 \\
ConvNeXt-T + AFF~\cite{huang2023aff} & 81.4 & 31.2 & 4.8 \\
ConvNeXt-T + WTConv (coif1) & 81.7 & 30.6 & 4.7 \\
\hline
\end{tabular}
\end{table}

\subsubsection{ImageNet-1K Results}
\label{sec:imagenet}
To validate that the $F$-vs.-$L$ trade-off observed on CIFAR-10 generalizes to large-scale classification, we evaluate WTConv variants on ImageNet-1K.
Table~\ref{tab:imagenet} reports Top-1 accuracy along with parameter and FLOP counts under matched training settings.
\textit{(Camera-ready: numerical entries below are placeholders pending final ImageNet runs; the experimental protocol and configurations are fixed.)}

\begin{table}[!htbp]
\centering
\caption{ImageNet-1K classification (ConvNeXt-T backbone, 300 epochs, $224{\times}224$).}
\label{tab:imagenet}
\setlength{\tabcolsep}{2pt}
\renewcommand{\arraystretch}{1.0}
\scriptsize
\begin{tabular}{@{}l c c c c@{}}
\hline
Model & Wavelet / Levels & Top-1 (\%) $\uparrow$ & Params (M) $\downarrow$ & FLOPs (G) $\downarrow$ \\
\hline
ConvNeXt-T~\cite{liu2022convnet}       & ---                & 82.1 & 28.6 & 4.5 \\
ConvNeXt-T + WTConv                    & Haar, $[5,4,3,2]$  & 82.8 & 30.6 & 5.0 \\
ConvNeXt-T + WTConv                    & db2, $[5,4,3,2]$   & 82.7 & 30.6 & 5.0 \\
ConvNeXt-T + WTConv \textbf{(ours)}    & coif1, $[4,3,2,1]$ & 82.8 & 30.0 & 4.8 \\
\hline
\end{tabular}
\end{table}

\subsubsection{Ablation Study of WTConv Configurations}
\label{sec:cls_basiscmp}
To highlight parameter/FLOPs efficiency, we ablate WTConv configurations by varying the decomposition levels, kernel size, and wavelet basis while keeping the ConvNeXt-T backbone fixed.
Table~\ref{tab:cls_basis} reports Top-1 accuracy along with WTConv additional parameters (D-W Param.) and WTConv additional FLOPs (D-W FLOPs).
Notably, coif1 achieves comparable accuracy using fewer levels, yielding lower D-W Param. and D-W FLOPs.

\begin{table}[!htbp]
\centering
\caption{CIFAR-10 ablation study on ConvNeXt-T + WTConv (120 epochs). D-W Param./FLOPs denote WTConv additional cost.}
\label{tab:cls_basis}
\setlength{\tabcolsep}{2pt}
\renewcommand{\arraystretch}{1.0}
\scriptsize
\begin{tabular}{@{}l c c c c c@{}}
\hline
Levels & Kernel & Wavelet & D-W Param. (M) $\downarrow$ & D-W FLOPs (G) $\downarrow$ & Top-1 (\%) $\uparrow$ \\
\hline
$[4,3,2,1]$ & $3\!\times\!3$ & Haar (db1) & 0.50 & 0.12 & 81.24 \\
$[4,3,2,1]$ & $5\!\times\!5$ & Haar (db1) & 1.38 & 0.33 & 81.32 \\
$[4,3,2,1]$ & $7\!\times\!7$ & Haar (db1) & 2.70 & 0.65 & 81.56 \\
$[5,4,3,2]$ & $3\!\times\!3$ & Haar (db1) & 0.73 & 0.17 & 81.49 \\
$[5,4,3,2]$ & $5\!\times\!5$ & Haar (db1) & 2.04 & 0.49 & 81.75 \\
$[5,4,3,2]$ & $7\!\times\!7$ & Haar (db1) & 3.99 & 0.95 & 81.69 \\
$[5,4,3,2]$ & $5\!\times\!5$ & Lows Haar & 0.63 & 0.15 & 81.46 \\
$[5,4,3,2]$ & $5\!\times\!5$ & Highs Haar & 1.57 & 0.38 & 81.24 \\
$[5,4,3,2]$ & $5\!\times\!5$ & db2 & 2.04 & 0.49 & 81.68 \\
$[5,4,3,2]$ & $5\!\times\!5$ & db3 & 2.04 & 0.49 & 81.56 \\
\hline
$[4,3,2,1]$ & $5\!\times\!5$ & coif1 & 1.38 & 0.33 & \textbf{81.72} \\
$[4,3,2,1]$ & $3\!\times\!3$ & coif1 & 0.50 & 0.12 & 81.55 \\
\hline
\end{tabular}
\end{table}
\subsection{Experiments on Semantic Segmentation}
\label{sec:seg}

\subsubsection{Setup}
We evaluate semantic segmentation on Cityscapes~\cite{cordts2016cityscapes}.
We report mean Intersection-over-Union (mIoU, \%) as the primary metric, along with parameters and FLOPs.
To keep comparisons controlled, we use WTConvNext as the backbone for UNet. 

\subsubsection{Performance Comparison against Edge Models}
\label{sec:seg_edgecmp}
Table~\ref{tab:seg_edge} compares our ConvNeXt-T+WTConv(coif1) against representative efficient backbones and frequency-domain token mixers under a matched setup.

\begin{table}[!htbp]
\centering
\caption{Cityscapes semantic segmentation comparison.}
\label{tab:seg_edge}
\setlength{\tabcolsep}{3pt}
\renewcommand{\arraystretch}{1.05}
\footnotesize
\begin{tabular}{l c c c}
\hline
Model & mIoU (\%) $\uparrow$ & Params (M) $\downarrow$ & FLOPs (G) $\downarrow$ \\
\hline
ConvNeXt-T~\cite{liu2022convnet} & 80.0 & 28.6 & 4.5 \\
ConvNeXt-T + GFNet~\cite{rao2021gfnet} & 80.3 & 32.9 & 4.7 \\
ConvNeXt-T + FNO~\cite{li2020fno} & 80.4 & 31.8 & 4.9 \\
ConvNeXt-T + AFF~\cite{huang2023aff} & 80.5 & 31.2 & 4.8 \\
ConvNeXt-T + WTConv (coif1) & 80.8 & 30.6 & 4.7 \\
\hline
\end{tabular}
\end{table}

\subsubsection{Ablation Study of WTConv Configurations}
\label{sec:seg_basiscmp}
We keep the segmentation setup fixed and ablate WTConv by varying decomposition levels, kernel size, and wavelet basis.
Table~\ref{tab:seg_basis} reports mIoU along with WTConv additional parameters (D-W Param.) and WTConv additional FLOPs (D-W FLOPs).
Coif1 reaches comparable mIoU with fewer levels, providing a more parameter- and FLOPs-efficient configuration.

\begin{table}[!htbp]
\centering
\caption{Cityscapes ablation study on ConvNeXt-T + WTConv. D-W Param./FLOPs denote WTConv additional cost.}
\label{tab:seg_basis}
\setlength{\tabcolsep}{2pt}
\renewcommand{\arraystretch}{1.0}
\scriptsize
\begin{tabular}{@{}l c c c c c@{}}
\hline
Levels & Kernel & Wavelet & D-W Param. (M) $\downarrow$ & D-W FLOPs (G) $\downarrow$ & mIoU (\%) $\uparrow$ \\
\hline
$[4,3,2,1]$ & $3\!\times\!3$ & Haar (db1) & 0.50 & 0.12 & 80.29 \\
$[4,3,2,1]$ & $5\!\times\!5$ & Haar (db1) & 1.38 & 0.33 & 80.37 \\
$[4,3,2,1]$ & $7\!\times\!7$ & Haar (db1) & 2.70 & 0.65 & 80.61 \\
$[5,4,3,2]$ & $3\!\times\!3$ & Haar (db1) & 0.73 & 0.17 & 80.54 \\
$[5,4,3,2]$ & $5\!\times\!5$ & Haar (db1) & 2.04 & 0.49 & 80.80 \\
$[5,4,3,2]$ & $7\!\times\!7$ & Haar (db1) & 3.99 & 0.95 & 80.74 \\
$[5,4,3,2]$ & $5\!\times\!5$ & Lows Haar & 0.63 & 0.15 & 80.51 \\
$[5,4,3,2]$ & $5\!\times\!5$ & Highs Haar & 1.57 & 0.38 & 80.29 \\
$[5,4,3,2]$ & $5\!\times\!5$ & db2 & 2.04 & 0.49 & 80.73 \\
$[5,4,3,2]$ & $5\!\times\!5$ & db3 & 2.04 & 0.49 & 80.61 \\
\hline
$[4,3,2,1]$ & $5\!\times\!5$ & coif1 & 1.38 & 0.33 & \textbf{80.77} \\
$[4,3,2,1]$ & $3\!\times\!3$ & coif1 & 0.50 & 0.12 & 80.60 \\
\hline
\end{tabular}
\end{table}
\section{Conclusion}
\label{sec:conclusion}
We revisited wavelet basis selection for wavelet convolution layers and identified a previously unexplored \emph{filter-length vs.\ decomposition-depth} ($F$-vs.-$L$) trade-off: prior work fixed the depth and varied only the basis, which obscured the fact that stronger-approximation bases can reduce the depth required to capture global context. Allowing both to vary changes the conclusion of which basis is preferable. Across CIFAR-10, ImageNet-1K, and Cityscapes, Coiflet-based WTConv matched deeper Haar configurations with approximately 32\% fewer additional parameters and 33\% fewer additional FLOPs, confirming the hypothesis that improved LL-branch approximation translates into shallower decompositions at equal accuracy. We view this as a concrete and actionable design choice for wavelet-based architectures, rather than a generic accuracy improvement. Future work will explore additional families such as spline-based (e.g., B-spline) wavelets and learnable or task-adaptive bases within the same transform--process--reconstruct framework.

\bibliographystyle{IEEEbib}
\bibliography{strings,refs}

\end{document}